\documentclass{article}

\usepackage[utf8]{inputenc}
\usepackage{float}
\usepackage{graphicx}
\usepackage{tikz}
\usepackage{times}
\usepackage{hyperref}
\usepackage{booktabs}

\usepackage[accepted]{wfvml2023}

\graphicspath{ {./images/} }

\usepackage{amsmath}
\usepackage{amssymb}
\usepackage{mathtools}
\usepackage{amsthm}
\usepackage{xcolor}

\theoremstyle{plain}

\theoremstyle{definition}

\theoremstyle{remark}

\wfvmltitlerunning{Catching Image Retrieval Generalization}

\begin{document}

\twocolumn[
\wfvmltitle{Catching Image Retrieval Generalization}

\begin{wfvmlauthorlist}
\wfvmlauthor{Maksim Zhdanov}{company}
\wfvmlauthor{Ivan Karpukhin}{company}
\end{wfvmlauthorlist}

\wfvmlaffiliation{company}{Tinkoff}

\wfvmlcorrespondingauthor{Maksim Zhdanov}{m.a.zhdanov@tinkoff.ru}
\wfvmlcorrespondingauthor{Ivan Karpukhin}{i.a.karpukhin@tinkoff.ru}

\wfvmlkeywords{Machine Learning, Metric Learning, Formal Verification}

\vskip 0.3in
]

\printAffiliationsAndNotice{}


\begin{abstract}


The concepts of overfitting and generalization are vital for evaluating machine learning models. In this work, we show that the popular Recall@K metric depends on the number of classes in the dataset, which limits its ability to estimate generalization. To fix this issue, we propose a new metric, which measures retrieval performance, and, unlike Recall@K, estimates generalization. We apply the proposed metric to popular image retrieval methods and provide new insights about deep metric learning generalization.
\end{abstract}


\section{Introduction}


Recall@K, also known as top-K accuracy, is one of the most popular evaluation metrics in retrieval and metric learning \cite{musgrave2020reality}. Recall@K is closely related to practical applications, where the retrieval error rate must be measured and minimized. However, the good metric for evaluating machine learning models must exhibit additional properties, necessary for measuring overfitting and generalization \cite{huai2019deep, roelofs2019measuring}. In this paper we show that Recall@K lacks the desired properties. In particular, we show that metric values largely depend on the number of classes in the dataset and can't be used for quality comparison among dataset splits, including differences between train and test sets.

To overcome the problems, related to Recall@K, we propose Grouped Recall@K, a simple alternative to Recall@K, which is invariant to the number of classes in the dataset and thus can be used for evaluating overfitting and generalization. We further demonstrate the importance of measuring these quantities by studying deep image retrieval \cite{roth2020revisiting}. In particular, we show that Grouped Recall@K can diagnose overfitting and underfitting, while traditional Recall@K can not.

Contributions of the paper can be summarized as follows:
\begin{itemize}
\item{We show Recall@K depends on the number of classes in the dataset, and thus can't be used for measuring overfitting and generalization.}
\item{We propose a simple Grouped Recall@K metric, that is invariant to the number of classes in the dataset and has a lower computational complexity, compared to Recall@K.}
\item{We show, that the proposed metric can measure the generalization gap. We also provide theoretical bounds for generalization on unseen data.}
\item{We apply the proposed metric to deep image retrieval and demonstrate the importance of measuring generalization and overfitting.}
\end{itemize}


\begin{figure}[t]  
\begin{center}
\centerline{\includegraphics[width=\columnwidth]{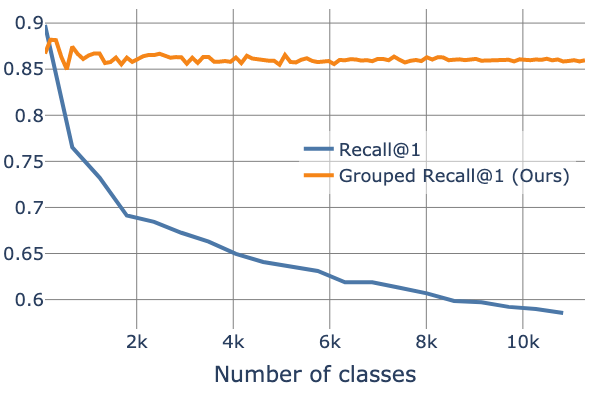}}
\caption{Illustration of the difference between the commonly used Recall@1 metric and the proposed Grouped Recall@1. Recall@1 reduces monotonically as the number of test classes increases. The proposed Grouped Recall@1 is invariant to the number of classes, but larger dataset sizes increase evaluation stability. Metrics are computed for the BN-Inception model on the Stanford Online Products (SOP) dataset.}
\label{figure_recall}
\end{center}
\end{figure}


\section{Motivation}
Generalization and overfitting were extensively studied in the context of classification. Suppose there is a dataset $\mathcal{X}$, sampled from the data distribution $\mathcal{D}$ and consisting of pairs $(x_i, y_i), i=\overline{1, N}$, where $x$ is a feature vector and $y$ is a class label. Suppose, there is a classification model $f(x)$. The classification accuracy of the model $f$ is defined as:
\begin{equation}
    \mathcal{A}(\mathcal{X}) = \frac{1}{N}\sum\limits_{i=1}^N \mathrm{I}(f(x_i) = y_i),
\label{eq:accuracy}
\end{equation}
where $\mathrm{I}$ is an indicator function. If pairs $(x_i, y_i)$ are sampled independently, then each element of the sum in Equation \ref{eq:accuracy} is independent of other elements, and several properties are held:
\begin{enumerate}
    \item{The expected value of accuracy is independent of the dataset size.}
    \item{The larger dataset leads to the more precise evaluation of the expected accuracy and theoretical bounds for expected accuracy can be estimated.}
\end{enumerate}

In particular, there are two important quantities for analyzing machine learning models. First is called {\it generalization gap} \cite{roelofs2019measuring}, and is evaluated as a difference of accuracy on the train set $\mathcal{X}_{train}$ and the test set $\mathcal{X}_{test}$:
\begin{equation}
    \mathcal{A}_{GAP} = \mathcal{A}(\mathcal{X}_{train}) - \mathcal{A}(\mathcal{X}_{test}).
\end{equation}
The second quantity is called {\it generalization error} and measures the difference between expected accuracy on the data distribution $\mathcal{D}$ and accuracy on the test set:
\begin{equation}
    \mathcal{A}_{Error} = \mathrm{E}_{(x, y) \in \mathcal{D}}\mathrm{I}(f(x) = y) - \mathcal{A}(\mathcal{X}_{train}).
\end{equation}
According to previous works, it is possible to estimate confidence intervals for $\mathcal{A}_{Error}$ with some confidence level $\alpha$ \cite{zhang2019intervals}.


In retrieval, we want to measure the ability of the model to find elements from the gallery dataset $\mathcal{D_G}$, that have the same class label $L(x)$, as a query element $q$. The retrieval model $f(q, \mathcal{D_G})$ maps the query and gallery to the index of the gallery element. If there is a set of query elements $Q$, then Recall@1 is defined as a misclassification rate:
\begin{equation}
\mathcal{R}@1(Q, \mathcal{D_G}) = \frac{1}{|Q|}\sum\limits_{i=1}^{|Q|}\mathrm{I}(L(x_{f(q_i, \mathcal{D_G})}) = L(q_i)).
\end{equation}
In practice, it is usually difficult to gather and annotate a sufficient amount of data for both queries $Q$ and gallery $\mathcal{D_G}$. An efficient way to use evaluation data is to collect a single labeled dataset $\mathcal{D}$, iterate over its elements, and consider each element as a query and the remaining part of the dataset as a gallery. Then Recall@1 can be evaluated as follows:
\begin{equation}
    \mathcal{R}@1(\mathcal{D}) = \frac{1}{|\mathcal{D}|}\sum\limits_{i=1}^{|\mathcal{D}|}\mathrm{I}(L(x_{f(q_i, \mathcal{D}_{/i}))} = L(x_i)),
    \label{eq:recall}
\end{equation}
where $\mathcal{D}_{/i}$ is a gallery, consisting of all elements, except $i$-th.

There are multiple problems related to the Recall@1 metric. One of them is that elements in the sum from Equation \ref{eq:recall} are dependent on each other. So the conditions of the central limit theorem are not met and theoretical bounds for generalization error can't be established. Another problem is that Recall@1 largely depends on the number of different labels in the dataset. An example of this dependency is presented in Figure \ref{figure_recall}. It can be seen, that the difficulty of the retrieval problem grows with the growing number of classes. So Recall@1 degrades on large datasets. Another problem relates to a wide range of models, based on the nearest neighbors search. These models search the gallery for the closest element to the query, leading to the total $O(N \log N)$ computational complexity. For large datasets, it is a limiting factor, which makes Recall@1 inapplicable.

In this paper, we raise the following question: how can we overcome the limitations of Recall@K, while preserving the informativeness of the metric? We answer this question by proposing a grouped alternative to Recall@K.

\section{Method}


The proposed Grouped Recall@K evaluation metric can be seen as a simple extension of the traditional Recall@K. Suppose we split all labels in the dataset into $N$ non-overlapping groups $G_i, i=\overline{1, N}$ with exactly $S$ labels in each. We then can estimate Recall@K on each group and average the results:
\begin{equation}
    \mathcal{GR}@K_S(\mathcal{D}) = \frac{1}{N}\sum\limits_{i=1}^{N}\mathcal{R}@K(\mathcal{D}_{G_i}),
\end{equation}
where $\mathcal{D}_{G_i}$ is a subset of $\mathcal{D}$ including all elements with labels from $G_i$.

\begin{table*}[t]
\caption{Comparison of the quality between two test sets (\%). }
\label{table_1}
\vskip 0.15in
\begin{center}
\begin{small}
\begin{sc}
\begin{tabular}{lccccc}
\toprule
Dataset & Recall@1 & Ours & Bound & Difference between two sets & Bound on difference \\ 
\midrule
CARS196 & 8.4 & 32.1 & 33.4 & 1.8 ± 0.8 & 2.8 ± 0.3 \\
CUB200 & 6.8 & 29.8 & 32.5 & 2.0 ± 1.0 & 6.0 ± 0.1 \\ 
InShop & 23.4 & 77.7 & 78.3 & 0.5 ± 0.2 & 1.3 ± 6$\times 10^{-4}$ \\  
SOP & 21.1 & 68.9 & 69.3 & 0.2 ± 0.1 & 0.9 ± 2$\times 10^{-4}$ \\  
\bottomrule
\end{tabular}
\end{sc}
\end{small}
\end{center}
\vskip -0.1in
\end{table*}
\subsection{Invariance to the number of classes}

By splitting the dataset into subsets with an equal number of labels, we equalize recall values on each portion. The scale of the grouped recall metric becomes dependent on the group size $K$ rather than on the number of classes in the dataset. This simplifies the comparison between different splits of the dataset, as we show in the experiments section.

\subsection{Generalization error bounds}
Since each group in Grouped Recall is independent of other groups, the Recall@K values on groups can be seen as i.i.d. random variables. We thus can apply the Central limit theorem. In particular, if the variance of metric values on groups is equal to $\sigma^2$, then the distribution of the Grouped Recall has the variance $\frac{\sigma^2}{r}$, where $r$ is the number of groups. We thus can find the required quantiles of the normal distribution, leading to the confidence interval.


\subsection{Computational complexity}

The basic KNN algorithm which is commonly used for image retrieval has $\mathcal{O}(n^2)$ computational complexity. However, some implementations utilize tree-based approaches like BallTree or KD-Tree \cite{scikit-learn} and allow to cut the complexity only to $\mathcal{O}(n\log{n})$. Having $r$ groups in total, the proposed metric achieves complexity $\mathcal{O}(n\log{\frac{n}{r}})$. The higher number of groups leads to faster computation. It scales \textbf{linearly} with dataset size. This observation is crucial for large-scale applications of retrieval systems.

\section{Related work}

Generalization has been a hot topic in research for a long time \cite{kawaguchi2017generalization, chatterjee2022generalization, jakubovitz2019generalization}.  Several studies have addressed the problems of generalization in image retrieval. For instance, \cite{bellet2015robustness} defines generalization bounds for image retrieval algorithms based on the notion of algorithmic robustness. Also, \cite{milbich2020sharing} aims to improve the generalization ability of metric learning methods via sharing characteristics between metric learning data. Generalization is referred to as an ability of a model to maintain performance on unseen data coming from some source domain \cite{maharaj2022generalizing}. In this definition, confidence intervals can be used to measure theoretical performance bounds on unseen data, providing an estimation of neural network generalization. 

Similarly, the concept of overfitting is widely studied in a deep learning community, and many approaches to tackle this problem were proposed: various penalties for model parameters \cite{bartlett2017spectrally, sedghi2018singular}, data augmentations \cite{yun2019cutmix, zhang2017mixup} and specific choice of optimization algorithms \cite{neyshabur2017implicit}. Overfitting has a strong connection to generalization and is often defined as a gap between train and test performance \cite{roelofs2019measuring}. However, it's not suitable for retrieval according to a simple observation (refer to Figure \ref{figure_recall}). While the previous statement holds, \cite{huai2019deep} highlights the importance of theoretical guarantees in the field which also can not be provided by traditional metrics. 

To the best of our knowledge, there is no metric for image retrieval tasks that can be used to estimate empirical generalization for test data and overfitting.

\section{Experiments}


\begin{table*}[t]
\caption{Overfitting, \%.}
\label{table_2}
\vskip 0.15in
\begin{center}
\begin{small}
\begin{sc}
\begin{tabular}{c|cccccc}
\toprule
Dropout & Train (ours) & Test (ours) & Train (recall) & Test (recall) & Gap (ours) & Gap (recall) \\ 
\midrule
0.0 & 68.5 & 66.0 & 44.0 & 41.1 & 2.5 & 2.9 \\
0.2 & 69.0 & 66.2 & 44.3 & 38.7 & 2.8 & 5.6 \\ 
0.5 & 69.6 & 66.0 & 45.0 & 36.1 & 3.6 & 8.9 \\  
0.7 & 70.2 & 65.5 & 45.9 & 38.1 & 4.7 & 7.8 \\ 
0.99 & 44.2 & 48.0 & 24.3 & 23.4 & \textbf{-3.8} & \textbf{0.9} \\
\bottomrule
\end{tabular}
\end{sc}
\end{small}
\end{center}
\vskip -0.1in
\end{table*}

In this section, we explore how our method can used for popular models and datasets in metric learning to study performance guarantees and overfitting.

\begin{figure}[t]
\vskip 0.2in
\begin{center}
\centerline{\includegraphics[width=\columnwidth]{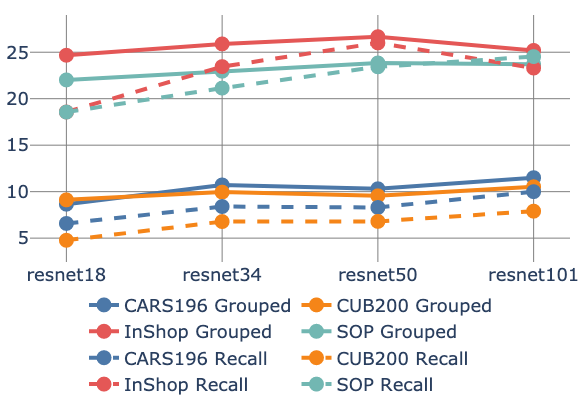}}
\caption{Comparison between test scores for Recall@1 and the grouped version. The proposed metric has high correlation with widely used Recall@1.}
\label{figure_comparison}
\end{center}
\vskip -0.2in
\end{figure}

\subsection{Generalization and bounds}

To empirically validate theoretical bounds on the generalization error, we compare the quality of the multiple methods on two subsets of the test set and measure the difference between metric values. For this purpose, we study four widely used benchmarks for image retrieval: CARS196 \cite{KrauseStarkDengFei-Fei_3DRR2013}, CUB200 \cite{WahCUB_200_2011}, InShop \cite{liuLQWTcvpr16DeepFashion}, SOP \cite{huai2019deep}. 
We show that the difference between qualities on test subsets lies within the theoretical bounds for Grouped Recall.
According to Table \ref{table_1}, Grouped Recall lies within its theoretical bounds, proving the ability of our metric to provide quality guarantees for test data.


\subsection{Relation to Recall@K}

According to the empirical results, presented in Figure \ref{figure_comparison}, the dynamics of the Grouped Recall metric are highly correlated with Recall@K. We thus conclude, that Grouped Recall measures the difference between the models, similar to Recall@K, and can be used for retrieval evaluation.

\subsection{Overfitting}
\label{overfitting}


To show that our metric can quantify overfitting via generalization gap we gradually add dropout to the ResNet34 \cite{he2016deep} feature extractor on the CIFAR100 dataset and we use the same splitting by 10 classes for a group. We examine that gap slowly approaches zero point for the proposed metric, whereas basic recall does not have such a feature.

\subsection{Model size}
\label{model_size}


\begin{figure}[t]
\vskip 0.2in
\begin{center}
\centerline{\includegraphics[width=\columnwidth]{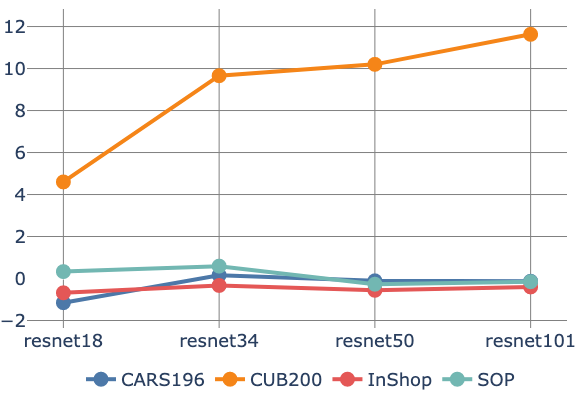}}
\caption{Model size relative to the generalization gap.}
\label{figure_1}
\end{center}
\vskip -0.2in
\end{figure}

In this experiment we study overfitting of the popular methods in deep image retrieval \cite{liu2017sphereface}. We do this by applying Grouped Recall and comparing generalization gaps of models with different sizes. 
According to Figure \ref{figure_1}, even large neural networks, such as ResNet101, lead to negative gaps on CARS196, InShop, and SOP datasets, which indicates the lack of overfitting. We thus suggest that deep networks are robust to overfitting on popular image retrieval datasets. This observation opens up new opportunities to study excessive neural network capacity for image retrieval.




\section{Conclusion}
In this work, we highlighted the limitations of the commonly used Recall@K metric in retrieval tasks. To overcome these limitations, we presented a simple approach to measuring retrieval performance with theoretical guarantees and the ability to quantify overfitting. Following the proposed evaluation methodology, we came to an unexpected observation, that common metric learning models demonstrate little overfitting. We thus suggest future work with a focus on studying large models in the context of image retrieval, even on medium and small datasets, like SOP, Inshop, and Cars196.

\bibliographystyle{wfvml2023}
\bibliography{ref}

\end{document}